\documentclass{article}
\usepackage{spconf}
\usepackage{cite}
\usepackage{amsmath,amssymb,amsfonts,amsthm}
\usepackage{bm}
\usepackage{graphicx}
\usepackage{url}
\usepackage{comment}
\usepackage{balance}

\newtheorem{definition}{Definition}
\newtheorem{proposition}{Proposition}
\newtheorem{theorem}{Theorem}
\newtheorem{lemma}{Lemma}

\DeclareMathOperator*{\argmax}{arg\,max}

\newcommand{\X}{\mathcal{X}}
\newcommand{\Y}{\mathcal{Y}}
\newcommand{\A}{\mathcal{A}}
\newcommand{\E}{\mathbb{E}}
\newcommand{\R}{\mathbb{R}}
\newcommand{\Prb}{\operatorname{Pr}}
\newcommand{\umax}{u_{\max}}
\newcommand{\OPT}{\mathrm{OPT}}

\begin{document}
\ninept
\linespread{1.045}\selectfont

\title{Risk-Averse Decision Making with Multi-Level Reliability Guarantees}

\name{Amirmohammad Farzaneh and Osvaldo Simeone}
\address{Institute for Intelligent Networked Systems (INSI), Northeastern University London, London, UK\\
\{a.farzaneh, o.simeone\}@nulondon.ac.uk}

\maketitle

\begin{abstract}
Many applications in engineering, including wireless broadcasting, require designs that provide performance certificates at different target outage levels. This paper studies the problem of maximizing the weighted average of such certificates in the presence of uncertainty about the true system state.
The problem is shown to be equivalent to an optimization over nested prediction sets, connecting to the literature on conformal prediction and extending prior art on single-level risk-averse decision making. Furthermore, we derive a dual formulation that decouples optimization across input values. Numerical experiments on a diversity-based wireless transmission system illustrate the cost of enforcing multi-level certificates with a single shared policy and trace the Pareto trade-off between multiple reliability levels.
\end{abstract}


%=============================================
\section{Introduction}
\label{sec:intro}
%=============================================

\begin{figure}[!t]
\centering
\includegraphics[width=\columnwidth]{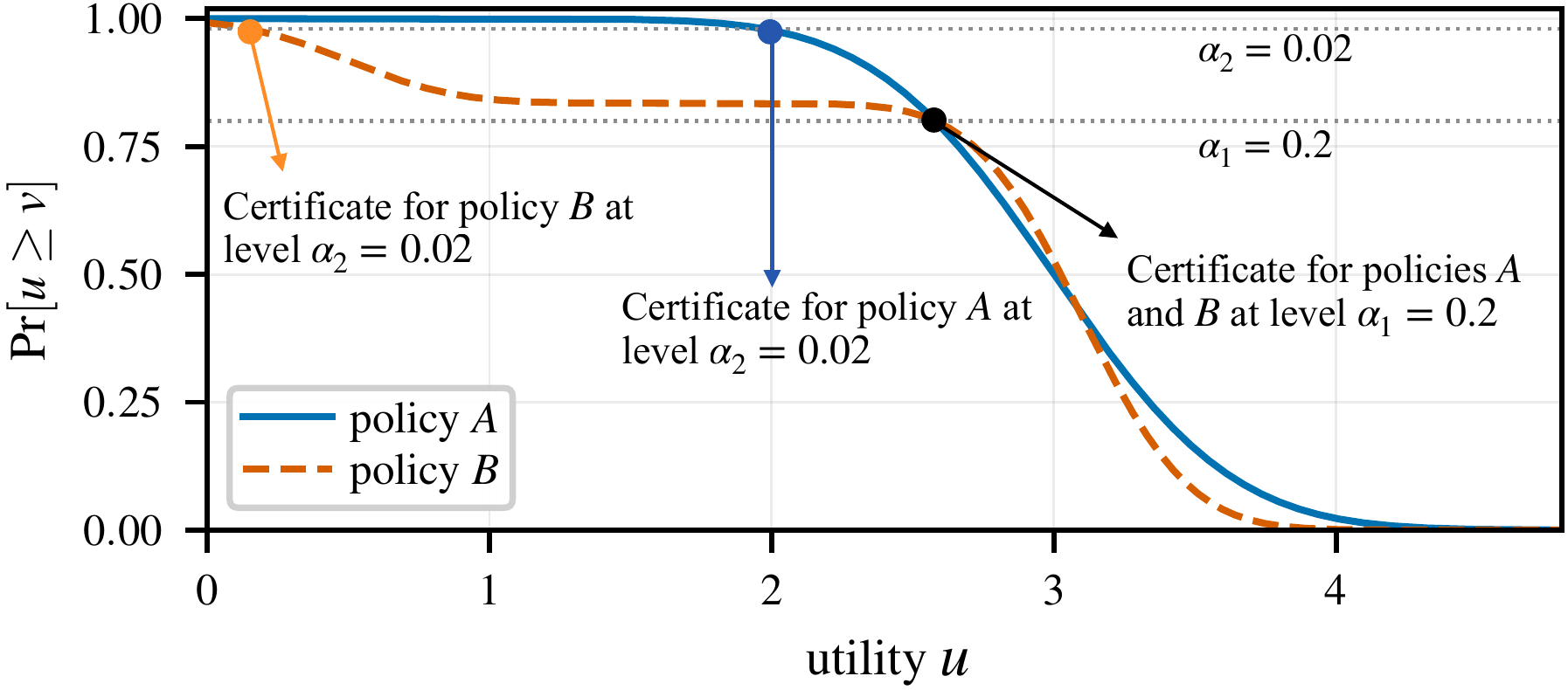}
\caption{Two utility distributions with matching utility certificates at outage level $\alpha_1=0.2$ but different utility certificates at the more stringent level $\alpha_2 = \alpha_1/10=0.02$.}
\label{fig:tailillu}
\end{figure}

Decisions under uncertainty require not only good nominal performance, but also predictable behavior as conditions deteriorate. For example, a wireless broadcast system may need to serve users whose connecting conditions range from line-of-sight to blocked propagation \cite{karasik2022learning}; a control system may face unexpected load spikes~\cite{astrom2008feedback}; and a learned classifier may encounter distributionally shifted inputs~\cite{quinonero2009dataset}. The standard formulation for optimal decision making includes an agent that observes features $X$, selects an action $a(X)$ without knowing the true state $Y$, and receives utility $u(a(X),Y)$. A risk-neutral policy maximizes expected utility, whereas a risk-averse policy also controls unfavorable outcomes, certifying a single utility level that is attained with probability at least $1-\alpha$ \cite{kiyani2025}. However, a certificate at a single outage level does not control the shape of the tail of the utility distribution. For example, as illustrated in Fig. \ref{fig:tailillu}, two policies that are indistinguishable at level $\alpha$ may behave very differently at a more stringent level, e.g., $\alpha/10$.

Taking inspiration from differentiated quality of service in telecommunications~\cite{wu2003effective,popovski2018urllc}, we study a single policy that supports $K$ utility certificates $\nu_1(X)\geq\cdots\geq\nu_K(X)$ at outage levels $\alpha_1\geq\cdots\geq\alpha_K$. Each pair $(\nu_k,\alpha_k)$ guarantees that utility $\nu_k(X)$ is attained in all but an $\alpha_k$-fraction of conditions. The ordering pairs stronger utility guarantees $\nu_k$ with more permissive outage levels $\alpha_k$ and weaker guarantees $\nu_k$ with more stringent reliability requirements $\alpha_k$. Together, these certificates can enforce a form of \emph{graceful degradation} in utility levels: the policy moves through progressively relaxed, but still certified, service levels as conditions worsen.

The single-level instance, i.e., $K = 1$, of the multi-level design problem described above was studied in \cite{kiyani2025}, where value-at-risk optimization is shown to lead naturally to prediction sets and max--min decision rules. The role of prediction sets in the problem provides a formal justification for the use of conformal prediction as an uncertainty quantification strategy \cite{vovk2005algorithmic, angelopoulos2023gentle}, as already proposed for a number of engineering applications \cite{simeone2026conformal, lindemann2025formal}. The framework in \cite{kiyani2025}, however, does not extend directly to a setting with $K>1$ levels, as applying it independently at each of the $K$ outage levels would generally produce $K$ different actions $\{a_k(X)\}_{k=1}^K$, and therefore would not define one deployable policy $a(X)$ with graded guarantees.

In Section~\ref{sec:formulation}, we formulate the multi-level risk-averse decision problem, and Section \ref{sec:dual} derives a dual characterization decoupling optimization across input values. In Section~\ref{sec:predset}, we establish an equivalent formulation based on $K$ nested prediction sets, extending the single-level equivalence of~\cite{kiyani2025} and providing a formal motivation for the use of nested conformal prediction~\cite{kuchibhotla2023nested, rivera2026online, ding2025calibrated}. Finally, Section~\ref{sec:experiments} uses a diversity-based wireless transmission example to illustrate the compromise between certificates at different service levels.

%=============================================
\section{Problem Formulation}
\label{sec:formulation}
%=============================================

Let $(X,Y)\sim P_{XY}$ denote the pair of agent's observation $X\in\X$ and unobserved state $Y\in\Y$. After observing $X=x$, the agent chooses an action $a(x)\in\A$ and receives utility $u(a(x),Y)\in[0,\umax]$ with $\umax<\infty$. Assuming $K$ service levels, fix maximum tolerated outage probabilities $\{\alpha_k\}_{k = 1}^K$ satisfying the inequalities
\begin{equation}
\label{eq:ineqalities}
1>\alpha_1\geq\cdots\geq\alpha_K>0,
\end{equation}
so that the required reliability $1-\alpha_k$ increases with the service level $k$. For each service level $k$, we wish to identify a utility certificate $\nu_k:\X\to[0,\umax]$ meeting the target reliability $1-\alpha_k$, i.e.,
\begin{equation}
\Prb\!\left[u(a(X),Y)\geq\nu_k(X)\right]\geq 1-\alpha_k,
\label{eq:outage}
\end{equation}
where the certificates naturally obey the pointwise ordering
\begin{equation}
\nu_1(x)\geq\nu_2(x)\geq\cdots\geq\nu_K(x)
\quad\text{for all }x\in\X
\label{eq:graceful}
\end{equation}
By condition~\eqref{eq:graceful}, larger utility certificates $\nu_k(x)$ are assigned to more permissive outage requirements $\alpha_k$.

\begin{definition}[RA-DPO$(\boldsymbol{\alpha})$]
For a vector of target outage probabilities $\boldsymbol{\alpha}=(\alpha_1,\ldots,\alpha_K)$ satisfying the inequalities \eqref{eq:ineqalities}, given a vector of non-negative weights $\{w_k\}_{k = 1}^K$ with $w_k\geq 0$ and $\sum_{k = 1}^K w_k = 1$, the $K$-level risk-averse decision policy optimization (RA-DPO) problem is defined as
\begin{equation}
\begin{aligned}
\max_{a(\cdot),\,\nu_1(\cdot),\ldots,\nu_K(\cdot)}
&\quad \sum_{k=1}^K w_k\E_X[\nu_k(X)]\\
\text{s.t.}
&\quad \Prb\!\left[u(a(X),Y)\geq\nu_k(X)\right]
\geq 1-\alpha_k,\\[-0.2em]
&\hspace{7.5em} k=1,\ldots,K,\\
&\quad \nu_1(x)\geq\nu_2(x)\geq\cdots\geq\nu_K(x),\\[-0.2em]
&\hspace{7.5em} x\in\X,
\end{aligned}
\label{eq:radpok}
\end{equation}
and its optimal value is denoted as $\OPT(\boldsymbol{\alpha})$.
\end{definition}

When $K=1$ and $w_1=1$, problem \eqref{eq:radpok} corresponds to the single-level RA-DPO problem studied in \cite{kiyani2025}. For $K>1$, the certificates $\{\nu_k(x)\}_{k = 1}^K$ cannot generally be optimized in isolation, since they all depend on the same action $a(x)$ that needs to cater to all service levels. The next basic result connects problem \eqref{eq:radpok} to its single-level counterpart, whose optimal value at outage level $\alpha$ is denoted as $\OPT(\alpha)$.

\begin{lemma}
\label{prop:sandwich}
The optimal value $\OPT(\boldsymbol{\alpha})$ for $\text{RA-DPO}(\boldsymbol{\alpha})$ can be bounded as
\begin{equation}
\OPT(\alpha_K)
\leq
\OPT(\boldsymbol{\alpha})
\leq
\sum_{k=1}^K w_k\OPT(\alpha_k),
\label{eq:sandwich}
\end{equation}
where $\OPT(\alpha)$ is the optimal value of $\text{RA-DPO}(\alpha)$ with $K = 1$.
\end{lemma}

The lower bound in \eqref{eq:sandwich} is the optimal value for the policy designed for the strictest reliability level $\alpha_K$ only. In contrast, the upper bound allows every service level $k$ to select its own action policy $a_k(X)$ and is therefore generally unattainable when one common policy $a(X)$ must serve all levels. The next sections elaborate on the optimization of problem \eqref{eq:radpok}.

\section{Dual Formulation}
\label{sec:dual}

Problem~\eqref{eq:reduced} jointly optimizes the $K+1$ functions $a(\cdot)$, $t_1(\cdot)$, $\ldots$, $t_K(\cdot)$, whose values at different inputs are coupled by the expected coverage constraints. In this section, we recast \eqref{eq:radpok} as a joint optimization over an action and a vector of local reliability allocations. We then show that this problem can be addressed via a dual formulation that decouples across values of $x$. To start, define
\begin{equation}
t_k(x)
=
\Prb\!\left[u(a(x),Y)\geq\nu_k(x)\mid X=x\right]
\label{eq:localcoverage}
\end{equation}
as the reliability at service level $k$ and input $x$, so that the constraint in \eqref{eq:outage} can be written as $\E_X[t_k(X)]\geq 1-\alpha_k$, and the constraint \eqref{eq:graceful} enforces the pointwise ordering $t_1(x)\leq\cdots\leq t_K(x)$. For a given allocation $t_k(x)$ of outage levels, the largest utility that can be certified for action $a$ is the conditional quantile
\begin{equation}
\begin{aligned}
Q_{t_k(x)}(x;a)
&=
\sup\bigl\{v\in[0,\umax]:\\
&\hspace{2.7em}
\Prb[u(a,Y)\geq v\mid X=x]\geq t_k(x)\bigr\}.
\end{aligned}
\label{eq:var_util}
\end{equation}
Substituting these expressions in the objective of \eqref{eq:radpok}, RA-DPO$(\boldsymbol{\alpha})$ can be equivalently stated as
\begin{equation}
\begin{aligned}
\max_{a(\cdot),\,t_1(\cdot),\ldots,t_K(\cdot)}
&\quad \sum_{k=1}^K w_k
\E_X\!\left[Q_{t_k(X)}(X;a(X))\right]\\
\text{s.t.}
&\quad \E_X[t_k(X)]\geq1-\alpha_k,\\[-0.2em]
&\hspace{7.5em} k=1,\ldots,K,\\
&\quad 0\leq t_1(x)\leq\cdots\leq t_K(x)\leq1,\\[-0.2em]
&\hspace{7.5em} x\in\X.
\end{aligned}
\label{eq:reduced}
\end{equation}

 Dualizing the constraints in \eqref{eq:reduced}, we now demonstrate that the maximization in \eqref{eq:reduced} can be solved independently at each input $x$. To see this, let
\begin{equation*}
\mathcal{T}=\{\mathbf{t}(x)\in[0,1]^K:
t_1(x)\leq\cdots\leq t_K(x)\},
\end{equation*}
so that the dual problem \cite{vandenberghe2004convex} for \eqref{eq:reduced} can be written as
\begin{equation}
\min_{\boldsymbol{\beta} \geq 0}
\left\{
\begin{aligned}
d(\boldsymbol{\beta})
&= \E_X\!\left[
\max_{\substack{a(X)\in\A\\\mathbf{t}(X)\in T}}
\left\{
\begin{aligned}
&\sum_{k=1}^K w_k Q_{t_k(X)}(X;a(X))\\[-0.2em]
&\quad+\sum_{k=1}^K\beta_k t_k(X)
\end{aligned}
\right\}
\right]\\
&\quad-\sum_{k=1}^K\beta_k(1-\alpha_k)
\end{aligned}
\right\}.
\label{eq:dual_problem}
\end{equation}
Generalizing \cite[Theorem 3.2]{kiyani2025}, solving problem \eqref{eq:dual_problem} is shown next to yield a solution also for problem \eqref{eq:reduced}.

\begin{theorem}
\label{thm:calibration}
Under the stated regularity conditions detailed in Appendix \ref{app:calibration}, the dual problem \eqref{eq:dual_problem} admits an optimal solution $\boldsymbol{\beta}^*\geq\mathbf{0}$, and an optimal solution for problem \eqref{eq:reduced} can be found separately for each input $x\in \mathcal{X}$ as
\begin{equation}
\begin{aligned}
(a^*(x),\mathbf{t}^*(x))
\in
\argmax_{\substack{a(x)\in\A\\\mathbf{t}(x)\in T}}
\biggl\{&
\sum_{k=1}^K w_kQ_{t_k(x)}(x;a(x))\\[-0.2em]
&{}+\sum_{k=1}^K\beta_k^*t_k(x)
\biggr\}.
\end{aligned}
\label{eq:jointargmax}
\end{equation}
\end{theorem}

%=============================================
\section{Prediction-Set Formulation}
\label{sec:predset}
%=============================================

In this section, we show that problem \eqref{eq:radpok} admits an equivalent formulation expressed in terms of $K$ nested prediction sets $C_k(x)\subseteq\Y$ with
\begin{equation}
    C_1(x)\subseteq\ldots\subseteq C_K(x)
\end{equation}
in the space of states $\Y$. This view connects RA-DPO to nested conformal prediction~\cite{kuchibhotla2023nested, rivera2026online, ding2025calibrated}, and gives the optimal decision rule produced by RA-DPO a worst-case, robust-optimization interpretation. Specifically, extending \cite[Theorem~2.3]{kiyani2025}, we have the following result.
\begin{proposition}
\label{prop:equivalence}
    RA-DPO$(\boldsymbol{\alpha})$ is equivalent to the problem
    \begin{equation}
\begin{aligned}
\max_{C_1(\cdot)\subseteq\cdots\subseteq C_K(\cdot)}
&\quad
\E_X\!\left[
\max_{a\in\A}
\sum_{k=1}^K w_k\inf_{y\in C_k(X)}u(a,y)
\right]\\
\text{s.t.}
&\quad
\Prb[Y\in C_k(X)]\geq1-\alpha_k,
\quad k=1,\ldots,K
\end{aligned}
\label{eq:racpo}
\end{equation}
in the sense that problems \eqref{eq:radpok} and \eqref{eq:racpo} have the same optimal value and optimal solutions for one problem yield optimal solutions for the other problem. Specifically, if $(a^*(x),\nu_1^*(x),\ldots,\nu_K^*(x))$ solves $\text{RA-DPO}(\boldsymbol{\alpha})$, then
\begin{equation}
C_k^*(x)
=
\{y\in\Y:u(a^*(x),y)\geq\nu_k^*(x)\}
\label{eq:nested}
\end{equation}
is optimal for \eqref{eq:racpo}. Conversely, if
$(C_1^*(x),\ldots,C_K^*(x))$ solves \eqref{eq:racpo}, then
\begin{align}
a^{*}(x)
&\in
\argmax_{a\in\A}
\sum_{k=1}^K w_k\inf_{y\in C_k^*(x)}u(a,y),
\label{eq:rule}\\
\nu_k^{*}(x)
&=
\inf_{y\in C_k^*(x)}u(a^{*}(x),y)
\label{eq:induced_certificate}
\end{align}
solve $\text{RA-DPO}(\boldsymbol{\alpha})$.
\end{proposition}

By \eqref{eq:nested}, each optimal prediction set is a utility superlevel set, and the ordering of certificates makes these sets nested. Moreover, the optimal policy \eqref{eq:rule} selects an action that maximizes the weighted sum of its worst-case utilities over all $K$ sets, with \eqref{eq:induced_certificate} providing the corresponding utility certificates. The proof is provided in Appendix~\ref{app:equivalence}.

%=============================================
\section{Numerical Example}
\label{sec:experiments}
%=============================================

We illustrate the impact of graceful-degradation requirements on a wireless communication problem consisting of a two-channel diversity transmission system~\cite{tse2005fundamentals,goldsmith2005wireless}. The first channel is characterized by a power gain $G_1\sim\operatorname{Exp}(1)$, corresponding to Rayleigh fading~\cite{goldsmith2005wireless}, while the second channel has power gain $G_2\sim\operatorname{Gamma}(6,1/6)$, corresponding to Nakagami-$6$ fading~\cite{goldsmith2005wireless} and is occasionally blocked. The blockage indicator $Z$ is partially known and distributed as $Z\mid X=x\sim\operatorname{Bern}(q(x))$ with $q(x)=0.85+0.05x$, where $X\sim\operatorname{Unif}[-1,1]$ is side information about link availability. The variables $X$, $G_1$, and $G_2$ are mutually independent, and the unobserved state is $Y=(Z,G_1,G_2)$. The action $a(x)\in[0,1]$ is the fraction of the power budget assigned to the second channel. Upon maximum ratio combining~\cite{tse2005fundamentals}, the combined channel gain is thus
\begin{equation}
H(a,Y)=(1-a)G_1+\kappa aZG_2,
\end{equation}
where $\kappa = 11\;\text{dB}$ reflects the higher nominal gain of the second channel, and the utility is given by the transmission rate
\begin{equation}
u(a,Y)=\min\bigl\{\log_2(1+\rho H(a,Y)),\umax\bigr\},
\end{equation}
with $\rho=10\;\text{dB}$ and $\umax=8$~bit/s/Hz. Overall, this setting models an always-available lower-gain channel, supplemented by a faster but blockage-prone channel.

Conditional on the blockage $Z$, the channel gain $H(a,Y)$ is either a scaled exponential ($Z=0$) or the sum of an exponential and an integer-shape Gamma variable ($Z=1$). Based on this, the conditional quantile $Q_t(x;a)$ can be obtained by monotone numerical inversion for any fixed pair $(x,a)$.
To solve the dual problem~\eqref{eq:dual_problem}, we approximate the continuous domains of $x$, $a$, and $t$ using uniform grids containing $2001$, $401$, and $201$ points, respectively, and pre-compute $Q_t(x;a)$ on the resulting grid. For each multiplier vector $(\beta_1,\beta_2)$, the pointwise maximization in \eqref{eq:dual_problem} is solved exactly on these discrete grids. The ordering $t_1\leq t_2$ is handled using a cumulative maximization over the admissible values of $t_1$, avoiding explicit enumeration of every pair $(t_1,t_2)$. The multipliers are obtained by coordinate bisection over finite intervals whose upper endpoints were verified to satisfy $\E_X[t_k^*(X)]\geq1-\alpha_k$.

\begin{figure}[!t]
\centering
\includegraphics[width=\columnwidth]{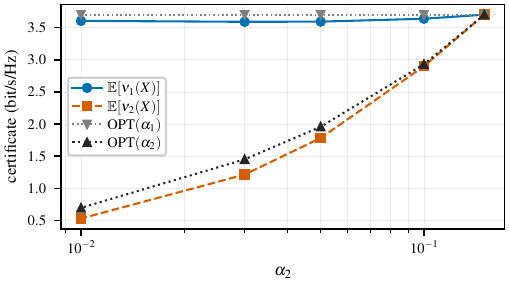}
\caption{Expected certificates $\mathbb{E}_X[\nu_1(X)]$ and $\mathbb{E}_X[\nu_2(X)]$ at equal weights $w_1=w_2$ for fixed outage probability $\alpha_1=0.15$ and varying $\alpha_2$. The logarithmic horizontal axis increases from the most stringent second-level requirement $\alpha_2=0.01$ on the left to the equal-outage case $\alpha_2=\alpha_1=0.15$ on the right. The dotted gray curve is the independently optimized single-level value $\OPT(\alpha_2)$}
\label{fig:equalw}
\end{figure}

We fix the first outage probability $\alpha_1=0.15$, corresponding to $85\%$ coverage, and vary the second outage probability $\alpha_2\in\{0.15,0.10,0.05,0.03,0.01\}$, corresponding to second-level coverages from $85\%$ to $99\%$. We first set equal weights $w_1=w_2$ to isolate the effect of tightening the second-level reliability requirement. As shown in Fig.~\ref{fig:equalw}, lowering the outage probability $\alpha_2$ tightens this requirement and therefore decreases the expected certificates $\E_X[\nu_1(X)]$ and $\E_X[\nu_2(X)]$. We also show the independently optimized values $\OPT(\alpha_1)$ and $\OPT(\alpha_2)$, quantifying the cost of using one action to support both reliability levels, rather than optimizing either certificate in isolation (see Lemma \ref{prop:sandwich}).

\begin{figure}[!t]
\centering
\includegraphics[width=\columnwidth]{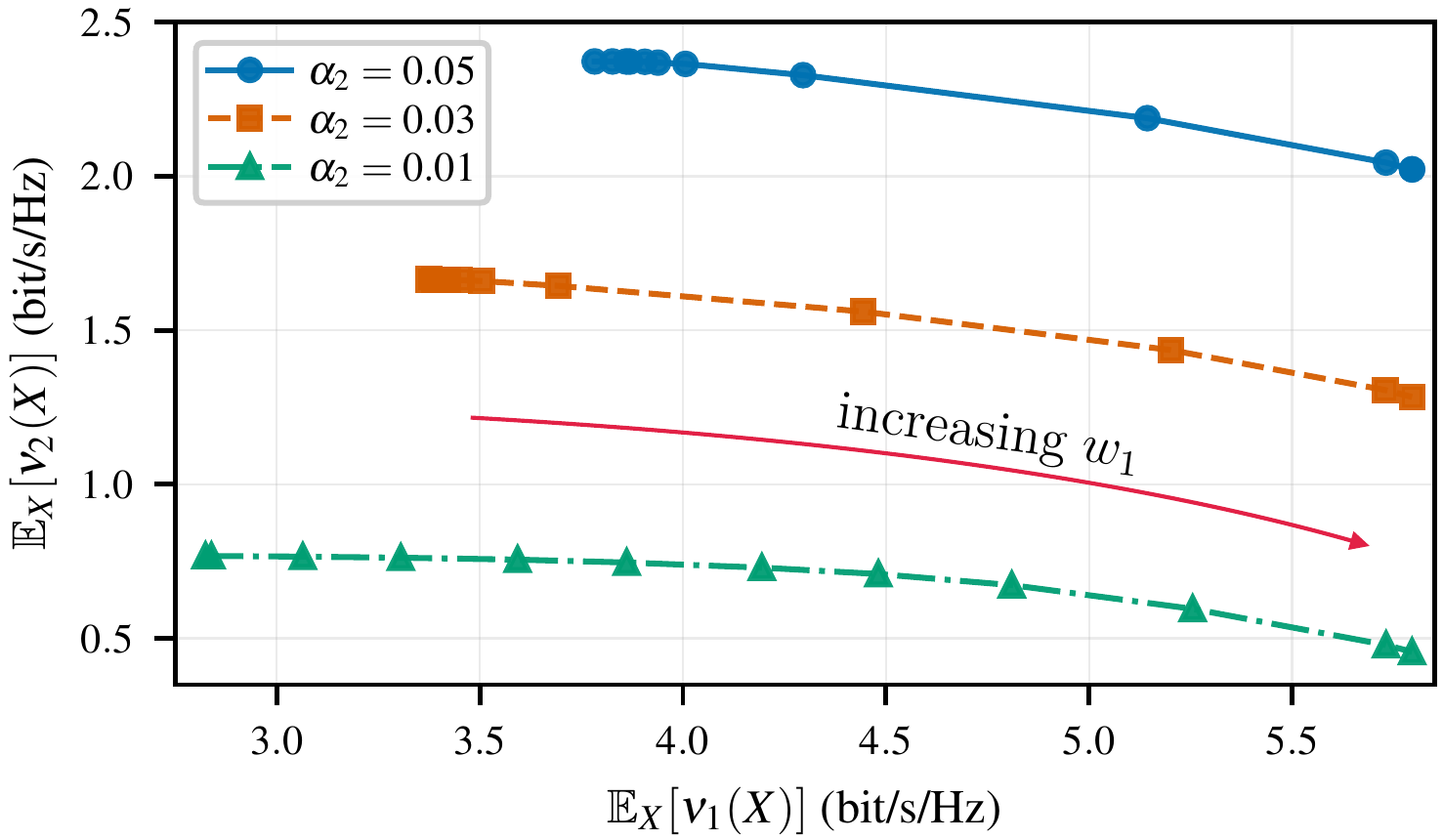}
\caption{Supported Pareto frontiers of the expected certificates for fixed outage probability $\alpha_1=0.20$ and selected values of $\alpha_2$. Along each frontier, the weight $w_1$ increases from left to right.}
\label{fig:pareto}
\end{figure}

Varying the weights $w_1$ and $w_2$ traces the trade-off between the two expected certificates $\mathbb{E}_X[\nu_1(X)]$ and $\mathbb{E}_X[\nu_2(X)]$. In Fig.~\ref{fig:pareto}, we fix the first outage probability $\alpha_1=0.2$, corresponding to $80\%$ coverage, and fix different values for the second outage probability $\alpha_2\in\{0.2, 0.15,0.10,0.05,0.03,0.01\}$. The figure highlights the price paid in terms of certificate $\mathbb{E}_X[\nu_1(X)]$ in order to increase the second level certificate $\mathbb{E}_X[\nu_2(X)]$. Moreover, decreasing the outage probability $\alpha_2$ shifts each frontier downward, since the second certificate must hold at higher reliability.

%=============================================
\section{Conclusion}
\label{sec:conclusion}
%=============================================

In this work, we have introduced a risk-averse decision framework in which a single action policy supports an ordered hierarchy of utility certificates. The formulation is shown to be equivalent to an optimization over nested prediction sets and admits a dual characterization with one scalar multiplier per service level. The resulting optimal policy lies between the strictest single-level solution and the weighted collection of independently optimized single-level solutions. A numerical example involving a diversity-based communication system illustrates how the shared action couples the two expected certificates. Developing more efficient methods for solving the pointwise problems, particularly for higher-dimensional applications, and constructing a distribution-free finite-sample calibration procedure for the full hierarchy are natural directions for future work.
%=============================================
\appendix
\section{Proof of Proposition 1}
\label{app:equivalence}
%=============================================

We prove equality of the optimal values in both directions. First, take a feasible RA-DPO solution $(a,\nu_1,\ldots,\nu_K)$ and define $C_k$ as in \eqref{eq:nested}. The certificate ordering makes these sets nested, and
\begin{equation}
\{Y\in C_k(X)\}
=
\{u(a(X),Y)\geq\nu_k(X)\},
\end{equation}
so every set satisfies the required marginal coverage. If $C_k(x)$ is nonempty, then
\begin{equation}
\inf_{y\in C_k(x)}u(a(x),y)\geq\nu_k(x).
\end{equation}
The same inequality holds for an empty set under our convention, since $\nu_k(x)\leq\umax$. Evaluating the objective in \eqref{eq:racpo} at the original action $a(x)$ therefore gives a value at least as large as the RA-DPO value. It follows that the RA-CPO optimum is no smaller than the RA-DPO optimum.

Conversely, take any feasible nested family $(C_1,\ldots,C_K)$ and define $a^*$ and $\nu_k^*$ by \eqref{eq:rule}--\eqref{eq:induced_certificate}. Since $C_k(x)\subseteq C_{k+1}(x)$, taking the infimum over the larger set cannot increase the result; hence the induced certificates are ordered. Furthermore,
\begin{equation}
\{Y\in C_k(X)\}
\subseteq
\{u(a^*(X),Y)\geq\nu_k^*(X)\}.
\end{equation}
Thus the induced policy and certificates are feasible for RA-DPO. Their RA-DPO objective is exactly the RA-CPO objective of the original sets. The RA-DPO optimum is therefore no smaller than the RA-CPO optimum. Combining the two inequalities proves equality, and applying the constructions to optimal solutions gives the stated correspondences.

%=============================================
\section{Proof of Theorem 1}
\label{app:calibration}
%=============================================

We proceed under the following regularity conditions: (i)~$\X$ is a standard Borel space and $P_X$ is non-atomic; (ii)~$\A\subset\R^d$ is compact; and (iii)~for $P_X$-almost every $x$, the mapping $(a,\mathbf{t})\mapsto\sum_k w_k Q_{t_k(x)}(x;a)$ is upper semicontinuous on $\A\times T$ and jointly measurable in $(x,a,\mathbf{t})$. The utility $u:\A\times\Y\to[0,\umax]$ is bounded as in Section~\ref{sec:formulation}.

The argument follows the convexify--dualize--recover proof of \cite[Theorem~3.2 and Appendix~A.4]{kiyani2025}, applied jointly to the ordered allocation vector. Write $G(x,a(x),\mathbf{t}(x))=\sum_k w_k Q_{t_k(x)}(x;a(x))$ and introduce the hypograph correspondence
\begin{equation*}
\begin{aligned}
\Gamma(x)=\bigcup_{a(x)\in\A}\bigl\{(\mathbf{t}(x),r):\;&\mathbf{t}(x)\in T,\\[-0.2em]
&0\leq r\leq G(x,a(x),\mathbf{t}(x))\bigr\}.
\end{aligned}
\end{equation*}
Joint upper semicontinuity and compactness of $\A\times T$ make $\Gamma(x)$ compact-valued, while the stated measurability conditions ensure measurable selections~\cite[Theorem~14.37]{rockafellar1998variational}. Its Aumann integral $\mathcal{S}=\int\Gamma(x)\,dP_X(x)$ is compact and, because $P_X$ is non-atomic, convex~\cite[Theorems~1, 3, and~4]{aumann1965integrals}. The reduced problem \eqref{eq:reduced} is therefore equivalent to maximizing $r$ over $(\mathbf{m},r)\in\mathcal{S}$ subject to $\mathbf{m}\geq\mathbf{1}-\boldsymbol{\alpha}$. Every policy gives such a point, and a measurable action realizing the upper boundary can be selected in the reverse direction. Thus including the hypograph only adds dominated objective values and does not change the optimum.

For any $\varepsilon\in(0,\min_k\alpha_k)$, the ordered constant allocation $t_k(x)=1-\alpha_k+\varepsilon$ is strictly feasible. Strong duality for this finite-dimensional convex problem therefore yields multipliers $\boldsymbol{\beta}^*\geq\mathbf{0}$, primal feasibility $\E_X[t_k^*(X)]\geq1-\alpha_k$, and the complementary-slackness conditions $\beta_k^*\bigl(\E_X[t_k^*(X)]-(1-\alpha_k)\bigr)=0$ for $k=1,\ldots,K$. Maximizing the corresponding Lagrangian over $\mathcal{S}$ is equivalent to maximizing its integrand pointwise, and a measurable maximizing selection exists by the same regularity conditions. Selecting an action that attains the upper boundary of $\Gamma(x)$ gives exactly \eqref{eq:jointargmax}. Finally, setting $\nu_k(x)= Q_{t_k^*(x)}(x;a^*(x))$ recovers the certificates that solve RA-DPO$(\boldsymbol{\alpha})$.

\begin{comment}
%=============================================
\section{Proof of Proposition 2}
\label{app:sandwich}
%=============================================

For the lower bound, take an optimal policy and certificate for the strictest level $\alpha_K$ and use the same certificate at every level. Since $\alpha_k\geq\alpha_K$, the strictest coverage guarantee implies every more permissive one. Equal certificates also satisfy the ordering constraint, so this construction is feasible for RA-DPO and has value $\OPT(\alpha_K)$.

For the upper bound, relax the shared-action and certificate-ordering constraints and optimize each service level independently. The relaxed objective separates across $k$ and has value $\sum_k w_k\OPT(\alpha_k)$. A relaxation cannot have a smaller optimum than the original problem, which proves the result.
\end{comment}

\vfill\pagebreak
%=============================================
\section*{Acknowledgments}
%=============================================
This work was supported by the European Research Council (ERC) under the European Union's Horizon Europe program (grant agreement No.~101198347). The work of O. Simeone was also supported by an EPSRC Open Fellowship (EP/W024101/1) and by the EPSRC project EP/X011852/1.

%=============================================
\balance
\bibliographystyle{IEEEbib}
\bibliography{bib}
%=============================================

\end{document}